# A Robust Deep Learning Workflow to Predict Multiphase Flow Behavior during Geological $CO_2$ Sequestration Injection and Post-Injection Periods

Bicheng Yan*, Bailian Chen, Dylan Robert Harp, Rajesh J. Pawar

Earth and Environmental Sciences, Los Alamos National Laboratory

*Corresponding author

Email: bichengyan@lanl.gov (B. Yan); dharp@lanl.gov (D.R. Harp); bailianchen@lanl.gov (B. Chen); rajesh@lanl.gov (R.J. Pawar).

**Abstract**

Simulation of multiphase flow in porous media is essential to manage the geologic $CO_2$ sequestration (GCS) process, and physics-based simulation approaches usually take prohibitively high computational cost due to the nonlinearity of the coupled physics. This paper contributes to the development and evaluation of a deep learning workflow that accurately and efficiently predicts the temporal-spatial evolution of pressure and $CO_2$ plumes during injection and post-injection periods of GCS operations. Based on a Fourier Neuron Operator, the deep learning workflow takes input variables or features including rock properties, well operational controls and time steps, and predicts the state variables of pressure and $CO_2$ saturation. To further improve the predictive fidelity, separate deep learning models are trained for $CO_2$ injection and post-injection periods due the difference in primary driving force of fluid flow and transport during these two phases. We also explore different combinations of features to predict the state variables. We use a realistic example of $CO_2$ injection and storage in a 3D heterogeneous saline aquifer, and apply the deep learning workflow that is trained from physics-based simulation data and emulate the physics process. Through this numerical experiment, we demonstrate that using two separate deep learning models to distinguish post-injection from injection period generates the most accurate prediction of pressure, and a single deep learning model of the whole GCS process including the cumulative injection volume of $CO_2$ as a deep learning feature, leads to the most accurate prediction of $CO_2$ saturation. For the post-injection period, it is key to use cumulative $CO_2$ injection volume to inform the deep learning models about the total carbon storage when predicting either pressure or saturation. The deep learning workflow not only provides high predictive fidelity across temporal and spatial scales, but also offers a speedup of 250 times compared to full physics reservoir simulation, and thus will be a significant predictive tool for engineers to manage the long term process of GCS.

**Keywords**: Deep learning, physics-based simulation, geologic $CO_2$ sequestration

## 1. Introduction

The process of geologic $CO_2$ sequestration (GCS) (Chen et al., 2018; Chen et al., 2020) can be described by the physics of multiphase flow and transport in porous media. In physics-based reservoir simulations, the governing partial differential equations of multiphase flow in porous media are discretized into a large scale algebraic system based on traditional numerical methods and iteratively solved to predict the temporal-spatial evolution of state variables (e.g., pressure and saturation) in porous media (Chen et al., 2006). Physics-based reservoir simulations are capable of handling many complex physics processes. However, as the nonlinearity of the processes grows due to heterogeneities in rock properties (Sohal et al., 2021), complex fluid thermodynamics (Li, et al., 2017; Michael et al., 2018), and coupled physics processes (Preisig and Prevost, 2011), the computational expense will be prohibitively high.

In recent years, deep learning (DL) techniques have had huge success to regress high dimensional data (Georgious et al., 2020) and approximate various functions (Csaji, 2001). In the area of fluid flow and transport in porous media, many research efforts have been put in improving the capability to predict the state variables in porous media with DL. For example, physics-informed neural network (PINN) models (Raissi et al., 2019) construct the loss function of the neural networks with the regularization of physics governing equations during the training process (Fuks and Tchelepi, 2020; Harp et al., 2021), such that the predictions by PINN are ensured to be consistent with the physics of fluid flow in porous media. The efficacy of PINN is demonstrated to predict processes governed by physics with low or medium complexity, but its computation may become intensive to solve flow problems with strong heterogeneity and high nonlinearity. On the other hand, image-based approaches have also been investigated to predict fluid flow in porous media, and mainly leverage convolutional neural networks (CNN) to approximate the nonlinear relationship between the input of rock property maps (e.g., permeability) and the output of fluid flow maps (e.g., saturation) in porous media (Zhong et al., 2019; Tang et al., 2020; Wen et al., 2021). With sparse connectivity between the input and output, image-based approaches tend to be more efficient at handling fluid flow problems with heterogeneous geological properties. During GCS, the multiphase flow physics controlling the evolution of pressure and saturation plumes is fairly complex in the $CO_2$ injection and post-injection periods, and thus the development of a general DL workflow to handle the whole GCS process becomes essential for efficient reservoir management of such processes.

Here we describe a DL workflow to predict the evolution of the state variables as multiple $CO_2$ injection and water production wells flow simultaneously in a 3D heterogeneous saline aquifer. This workflow is an image-based approach, as it takes full advantage of the spatial topology predictive capability of a Fourier Neural Operator (FNO) (Li et al., 2020). We mainly contribute to two research aspects in this work. First, we strategically develop and train exclusive DL models for different periods of the GCS process to improve the predictive accuracy of pressure based on the fact that in different periods the primary driving forces of fluid flow may vary. For example, the primary driving force during the injection period is the viscous force, while during the post-injection period, they are gravity and capillary pressure (if considered) for fluid re-equilibrium (Ide et al., 2007; Lee et al., 2016; Ren, 2018). Secondly, we also evaluate the impact of different combinations of input variables or features on the predictive accuracy of DL models, and find it is essential to treat the cumulative $CO_2$ injection volume as a DL feature in order to accurately predict the dynamic evolution of state variables, especially saturation.

The rest of the paper is organized as follows. In Section 2, we briefly review the physics of multiphase flow in porous media that governs the GCS process. In Section 3 we illustrate the DL workflow to emulate the GCS process in detail. In Section 4, the workflow is tested through numerical experiments to predict the migration of pressure and $CO_2$ plumes in a 3D heterogeneous saline aquifer, and the performance is comprehensively assessed. In Section 5, we conclude the work with a few remarks.

## 2. Governing Physics in GCS

As $CO_2$ is injected into a saline aquifer during the GCS process, the mobile fluid phases consist of a water-rich phase and a $CO_2$-rich phase, with water and $CO_2$ as the primary components in the water-rich and $CO_2$-rich phases, respectively. The flow and transport of each fluid component is governed by their corresponding mass balance equation,

$$\frac{\partial}{\partial t}\left(\phi \sum_\alpha S_\alpha \rho_\alpha x_{\alpha,i}\right) - \nabla \cdot \left\{K \sum_\alpha \frac{k_{r\alpha}}{\mu_\alpha} \rho_\alpha x_{\alpha,i} (\nabla p_\alpha + \rho_\alpha g \nabla Z)\right\} + \sum_l \left(\sum_\alpha \rho_\alpha x_{\alpha,i} q_\alpha\right)^l = 0, \qquad (1)$$

where the first term is the fluid accumulation, the second is the advective flux based on Darcy's law, and the third is the source or sink term. Subscript $i$ denotes the primary fluid components, including water and

$CO_2$; $\alpha$ denotes the fluid phase, including water-rich phase $w$ and $CO_2$-rich phase $g$; $t$ is time; $\phi$ is the rock porosity; $S_\alpha$ is the phase saturation; $\rho_\alpha$ is the fluid phase density; $x_{\alpha,i}$ is the mole fraction of component $i$ in fluid phase $\alpha$; $K$ is the rock permeability; $k_{r\alpha}$ is the phase relative permeability; $\mu_\alpha$ is the phase viscosity; $p_\alpha$ is the phase pressure; $g$ is the acceleration due to gravity; $Z$ is depth; and $q_\alpha$ denotes the rate for producing or injecting fluid phase $\alpha$ through well perforation $l$. Further, Equation (1) is constrained by several auxiliary relationships, including the equilibrium between fluid volume and pore space, capillary pressure constraint and fluid thermodynamics equilibrium. More details can be found in previous literature (Chen et al., 2006; Michael et al., 2018). In a physics-based reservoir simulator, these equations are solved iteratively to calculate the state variables of pressure and saturation.

## 3. Methodology of the Deep Learning (DL) Workflow

### 3.1 Feature Selection and Assembly

Features are defined as the input variables of the deep neural network (DNN) in the DL workflow, and the candidate features available consist of rock permeability $K$, rock porosity $\phi$, $CO_2$ injection rate $q_{inj}$, $CO_2$ cumulative injection volume $Q_{inj}$, locations of water injection wells $Loc_{prod}$ and time steps $t$. The state variables predicted by the DNN include pressure $p$ and saturation of the $CO_2$-rich phase $S_g$. Hereafter we shorten "saturation of $CO_2$-rich phase" to "saturation". These variables are consistently assembled as 2D images, as we adopt 2D convolutional neural network architecture for the DNN.

Among the static geological properties, the rock permeability $K$ is closely related to the spatial pore connectivity and contributes to the advective flux term in Equation (1). The permeability of geologic porous media is usually anisotropic, which means the vertical permeability $K_V$ is much lower than the horizontal permeability $K_H$, e.g., we assume $\frac{K_V}{K_H} = 0.1$ in our example. As a result, the impact of vertical connectivity by $K_V$ on fluid flow is negligible compared to that of horizontal connectivity by $K_H$. Therefore, ignoring the impact of vertical connectivity, we slice the 3D permeability volume into 2D horizontal layer-wise images. The rock porosity $\phi$ is a measure of void space of the rock and is only related to the fluid accumulation in Equation (1), and thus it can be directly sliced into 2D horizontal layer-wise images.

The $CO_2$ injection wells are considered to be operated following specified rate schedules, so two candidate features can honor these wells, including $CO_2$ injection rate $q_{inj}$ and $CO_2$ cumulative injection volume $Q_{inj}$, both of which are functions of time $t$. During the injection period, $q_{inj}$ and $Q_{inj}$ are representative variables for injection well controls. However, during post-injection period these wells are shut-in with no injection ($q_{inj} = 0$), and thus $Q_{inj}$ is always a constant equal to the ultimate $CO_2$ injection volume at the end of the injection period, which can be used as an important variable to honor the total $CO_2$ storage volume in the post-injection period. $q_{inj}$ and $Q_{inj}$ are assembled as 2D images with nonzero values in well perforated grid cells and zeros elsewhere. The water production wells are essential for the management of reservoir pressures during GCS, and in this work we treat the locations of these wells, $Loc_{prod}$, as a feature for generality. $Loc_{prod}$ is a 2D binary image with value 1 in the well perforated grid cells and 0 elsewhere. Additionally, a feature image filled with time $t$ is considered for temporal evolution of the state variables, and aids in interpolation at intermediate time steps without training data points.

As no capillary pressure is considered between the two primary phases in this work, the driving forces contributing to the advective flux term in Equation (1) are the viscous force and gravity. During the injection period, the primary driving force of fluid flow is the viscous force when the injection and production wells are open. However, as the injection and production wells are always shut-in in the post-injection period, the pressure and saturation fields start re-equilibrium, mainly driven by gravity. Hence, given the difference in primary driving forces of fluid flow and transport in different periods, we believe it also makes sense to develop separate and exclusive DNN models for the injection and post-injection periods.

In **Table 1**, we permutate the two periods of GCS and the candidate features, and generate 5 different DNN models to predict the state variables $p$ and $S_g$ separately, and we will evaluate the performance of each scenario through our numerical experiments.

**Table 1. Different DNN models used in this work**

| Scenarios | Periods | Feature Array $X$ | State Variable $Y$ |
|---|---|---|---|
| 1 | Injection | $[K, \phi, Loc_{prod}, q_{inj}, t]^T$ | $p$ or $S_g$ |
| 2 | Injection | $[K, \phi, Loc_{prod}, Q_{inj}, t]^T$ | $p$ or $S_g$ |
| 3 | Post-Injection | $[K, \phi, Loc_{prod}, Q_{inj}, t]^T$ | $p$ or $S_g$ |
| 4 | Injection & Post-Injection | $[K, \phi, Loc_{prod}, q_{inj}, t]^T$ | $p$ or $S_g$ |
| 5 | Injection & Post-Injection | $[K, \phi, Loc_{prod}, Q_{inj}, t]^T$ | $p$ or $S_g$ |

### 3.2 Deep Neural Network (DNN) Architecture

After the feature array $X$ and state variable vector $Y$ in **Table 1** are assembled, they are fed into the DNN model for training and inference processes. The DNN model we applied in this work is the Fourier Neural Operator (FNO) (Li et al., 2020), which directly operates on 2D images and has good predictive capability for different physics-based processes. The feature array $X$ is initially transformed into a high dimensional space $V_0$ through the first fully connected layer (FC-1), and then $V_l$ is iteratively updated through,

$$V_l = \sigma(WV_{l-1} + \kappa(V_{l-1})), l = 1, \dots, L. \tag{2}$$

where $V_l$ is the feature map at layer $l$, and is a function of $V_{l-1}$ preceding it; $\sigma$ denotes the nonlinear activation function; $W$ is a linear operator defined by a 1D convolutional operator; $\kappa$ is a 2D convolution operator defined in the Fourier space. In total we have 4 layers related to the operation in Equation (2), namely "Fourier" layers. Ultimately, $V_L$ is transformed back to the state variable $p$ or $S_g$ through two fully connected layers (FC-2 and FC-3). The architecture of FNO is depicted in **Fig. 1**. At FC-1, FNO takes the 5 feature images in $X$ (**Table 1**), then sequentially goes through 4 Fourier and 2 FC layers to predict the output image of state variable $p$ or $S_g$.

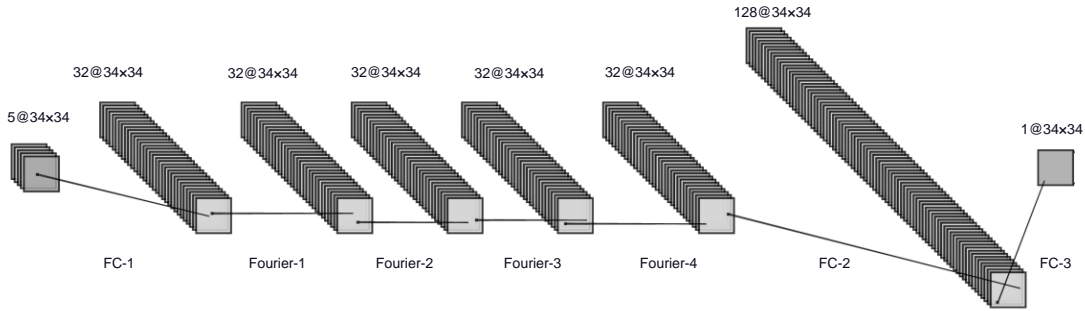

**Fig. 1.** Architecture of Fourier Neural Operator with 3 fully connected (FC) and 4 Fourier layers. "$a@n_x \times n_y$" at the top of each layer denotes: $a$ - number of features; $n_x$ - the image width; $n_y$ - the image length.

In **Fig. 1**, there is no activation function $\sigma$ in the input and output layers (FC-1 and FC-3), and $\sigma$ used through Fourier-1 to Fourier-4 is the LeakyReLU,

$$\text{LeakyReLU} = \begin{cases} x & if\ x \geq 0 \\ 0.01 & otherwise \end{cases}, \quad (3)$$

$\sigma$ used in FC-2 is ReLU,

$$\text{ReLU} = \begin{cases} x & if\ x \geq 0 \\ 0 & otherwise \end{cases}, \quad (4)$$

The loss function $\mathcal{L}$ in our FNO approach is defined as,

$$\mathcal{L}(\theta) = \|Y - \hat{Y}\|, \quad (5)$$

where $\theta$ are the learnable parameters; $\|\cdot\|$ is the root-mean-square-error operator; $Y$ is the ground truth of the pressure $p$ or saturation $S_g$; $\hat{Y}$ is the prediction of $Y$ by FNO. The ultimate goal of training FNO is to find $\theta$ by minimizing the loss function $\mathcal{L}$. We implemented FNO and the associated modules with the deep learning library PyTorch (Paszke et al., 2019), and adopted the Adam optimizer to train FNO.

The DL workflow of deep learning is shown in **Fig. 2**. The well related feature images, including $q_{inj}$, $Q_{inj}$ and $Loc_{prod}$, are full of zeros (blue pixels) except the locations (red pixels) where the injection or production wells are perforated. As the whole workflow is operated on 2D horizontal layer-wise images in the reservoir domain, we iteratively predict the state variables for each layer.

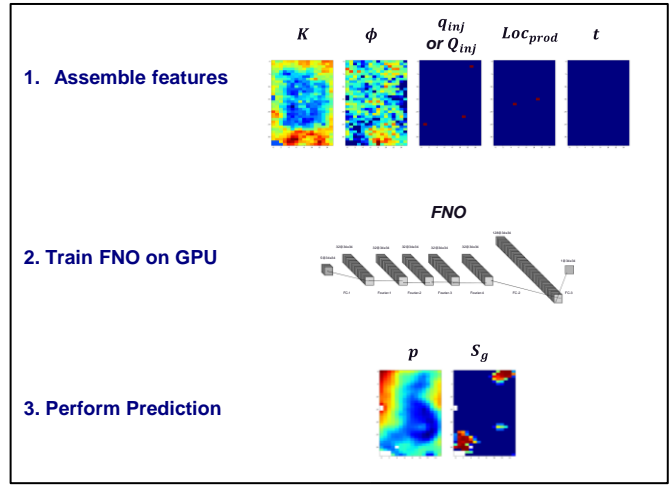

**Fig. 2.** DL workflow to learn and predict the pressure and saturation fields in GCS.

## 4. Numerical Experiments

### 4.1 Heterogeneous 3D Reservoir Model

In this section, a physics-based reservoir model is developed using a commercial reservoir simulator, GEM by CMG (CMG, 2020), which simulates the process of $CO_2$ injection and water production in a 3D heterogeneous saline aquifer. We use this model to generate synthetic data to train and test the DL workflow aforementioned. The reservoir domain is discretized using corner point grid, with $36 \times 16 \times 25$ grid cells in the $x$, $y$, $z$ directions, respectively. The permeability and porosity fields are correlated, and assumed to be spatially heterogeneous and uncertain. Multiple equiprobable realizations of heterogeneous permeability and porosity fields are generated using a geostatistical simulation approach, and ultimately 3 typical realizations at P10, P50 and P90 are selected. A P50 realization of the heterogeneous fields is depicted in **Fig. 3**, and there are 3 vertical wells for $CO_2$ injection (red wells) and 2 vertical wells for water production (green wells), all of which are perforated through all the reservoir layers. The total simulation life is 80 years, including 30 years with continuous $CO_2$ injection and water production, and the following 50 years of post-injection period with all wells shut-in. The simulation output is in monthly resolution, with 962 effective time steps in total.

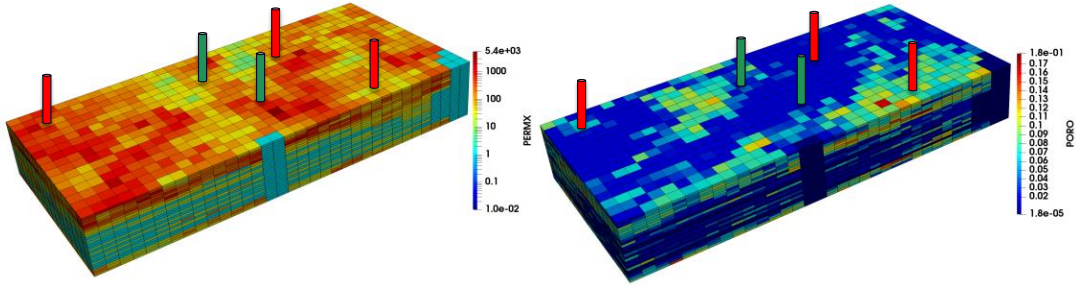

**Fig. 3. Permeability (left) and porosity (right) fields of the P50 realization. Red cylinder: $CO_2$ injection wells, green cylinder: water production wells.**

A total of 90 simulations were performed using different combinations of the 3 correlated permeability-porosity fields, total $CO_2$ injection mass, number of active injection wells and $CO_2$ injection mass split for active injection wells, as shown in **Table 2**. We choose 72 simulations for training (80%), 9 for validation (10%) and 9 for testing (10%).

**Table 2. Sampling parameters for the reservoir model**

| Parameters | Sample Values |
|---|---|
| Permeability $K$ realizations | P10, P50, P90 |
| Porosity $\phi$ realizations | P10, P50, P90 |
| Total $CO_2$ injection mass (tons) | 22.4, 11.2, 5.6 |
| Number of active injection wells | 2 or 3 |
| $CO_2$ injection mass split | Scenario of 2 active injection wells:<br>• 50/50 split<br>• 10/90 split<br>• 90/10 split<br>Scenario of 3 active injection wells:<br>• 33/33/33 split |

### 4.2 Model Training Performance

We assemble the samples for the DNN models following the scheme in Section 3.1. If we select the training and validation samples with the monthly resolution in the reservoir simulation output, the total number of these samples is 1,948,050 (962 time steps), which leads to potential computer memory issue and low training efficiency. To resolve this, we assemble these training and validation samples with yearly resolution, except that during the 1st year injection and 1st year of post-injection we assemble the samples with monthly resolution in order to accurately honor the early transient flow. Therefore, the total number of training and validation samples is reduced to be 210,600 (104 time steps in 81 training and validation simulations), with a relative reduction of 89%. As for the testing samples, we still assemble them in the original monthly resolution in order to perform rigorous comparison with reservoir simulations, and the total number of testing samples is 216,450 (962 time steps in 9 testing simulations).

For all the 5 DNN scenarios listed in **Table 1**, we select their training, validation and testing samples based on their corresponding prediction period(s). The training and validation samples are split into small batches with 20 samples/batch, the maximum number of training epochs is 20, and the learning rate is 0.001. We perform the training and prediction of the DNN models on a single GPU (NVIDIA Quadro RTX 4000).

**Table 3** illustrates the training CPU time and overall root-mean-square-error (RMSE) of pressure and saturation at the injection and post-injection periods (RMSE calculated based on testing samples only). The training CPU time is different among different scenarios, as the total number of training and validation samples of each scenario varies depending on their prediction period(s). Overall, the training process is very efficient. Further, the total training time for each scenario can be further improved by parallel processing since the pressure and saturation models are independent and can be trained concurrently. The pressure DNN model in the Scenario 1 has the lowest pressure RMSE (5.033 psia) in injection period, and the pressure DNN models in the Scenarios 3 and 5 leads to the two lowest pressure RMSE in post-injection period. The saturation DNN model in the Scenario 5 has the lowest overall saturation RMSEs in the injection period (0.01462) and post-injection period (0.03013). The Scenario 4 always have the highest RMSEs of both pressure and saturation, which can be explained by considering that the feature of $CO_2$ injection rate $q_{inj}$ used in this scenario is not a good feature in the post-injection period.

**Table 3. Training CPU time and overall RMSE of pressure and saturation at different periods**

| Scenario | Pressure Training CPU, hours | Saturation Training CPU, hours | Pressure RMSE, psia | | Saturation RMSE, fraction | |
|---|---|---|---|---|---|---|
| | | | Injection | Post-Injection | Injection | Post-Injection |
| 1 | 1.40 | 1.50 | <u>5.033</u> | NA | 0.01738 | NA |
| 2 | 1.48 | 1.62 | 7.611 | NA | 0.01990 | NA |
| 3 | 2.17 | 2.20 | NA | <u>2.767</u> | NA | 0.03372 |
| 4 | 3.67 | 3.60 | 8.440 | 18.461 | 0.02087 | 0.1267 |
| 5 | 3.57 | 3.54 | 5.965 | <u>2.732</u> | <u>0.01462</u> | <u>0.03013</u> |

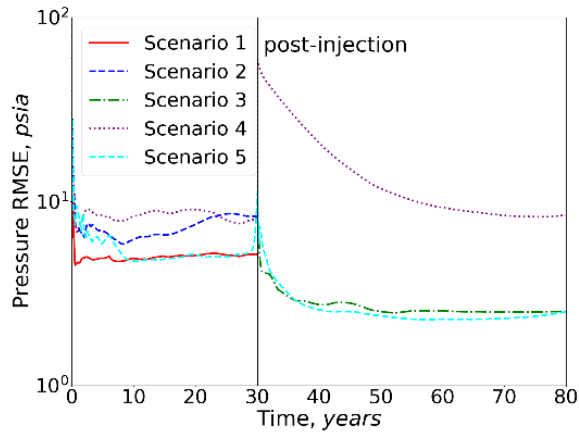

**Fig. 4. Pressure RMSE versus time for the 5 DNN scenarios in Table 1.**

Further, we calculate the pressure and saturation RMSEs versus time for different scenarios based on the 9 testing simulations. **Fig. 4** shows the dynamic change of pressure RMSE with time. At the beginning of injection (year 1), the pressure field in the reservoir is perturbed due to the instantaneous opening of the wells and almost all scenarios suffer the spikes of relatively high pressure RMSEs. Through comparison, Scenario 1 predicts the lowest overall pressure RMSE (red solid line) in the injection period, and it also exhibits the lowest RMSE (6.35 psia) in the 1st year. Similarly, the pressure field is also perturbed due to the instantaneous shut-in of the wells at the beginning of post-injection (year 30). Scenario 4 (purple dotted line) and Scenario 5 (cyan dashed line) suffers substantial discontinuity of pressure RMSE at around year 30. From the standpoint of feature distribution, the injection related features ($q_{inj}$ or $Q_{inj}$) used in these two scenarios are constant during the post-injection period due to well shut-in, while $q_{inj}$ and $Q_{inj}$ may vary with time during the injection period (well open) and thus the difference in feature distribution induces the occurrence of pressure discontinuity. From the intrinsic physics aspect, the significant pressure discontinuity in these two scenarios is because the primary driving force of fluid flow related to global pressure distribution is viscous force in the injection period, while that in the post-injection period becomes gravity. Whereas Scenario 3 (green dash-dot line) exclusively predicting post-injection period has the lowest pressure RMSE (6.47 psia) at the beginning of post-injection period, and its performance during the whole post-injection period is also highly faithful. Therefore, it necessitates to train separate DNN models for the injection period and the post-injection period in order to accurately predict pressure dynamics, and we ultimately choose to predict pressure $p$ through a combination of Scenarios 1 and 3, and the overall RMSE is 3.780 psia.

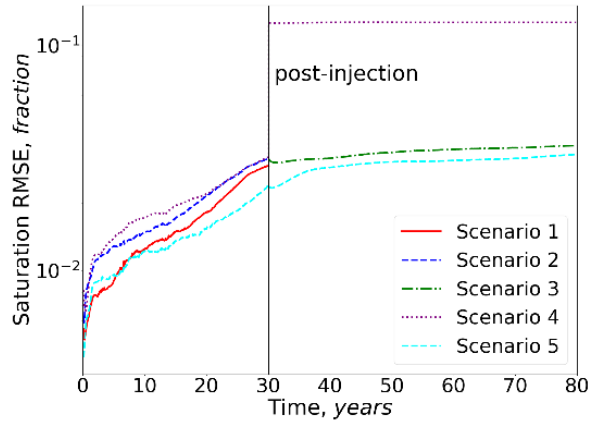

(b) Saturation RMSE

**Fig. 5. Saturation RMSE versus time for the 5 DNN scenarios in Table 1.**

The dynamic change of saturation RMSE with time is presented in **Fig. 5**. In the Scenarios 3, 4, and 5 that predict the post-injection period, Scenario 4 (purple dotted line) which treats $q_{inj}$ as a feature variable completely fails inferring the saturation in the post-injection period, because $q_{inj}$ is not an effective feature to inform the DNN model about how much $CO_2$ was ultimately injected into the reservoir after 30 years. Nevertheless, there is no significant discontinuity at the beginning of the post-injection (year 30) in Scenario 5 (cyan dashed line), which can be explained by considering that the saturation field responds slower to well shut-in than the pressure field. Scenario 5 brings the lowest saturation RMSE in both injection and post-injection periods, with overall RMSE 0.02542, and thus we choose it to predict the saturation $S_g$.

Based on the most effective scenarios to predict pressure (by Scenarios 1 and 3) and saturation (by Scenario 5), we further plot the RMSE of pressure and saturation in each reservoir layer, aggregated through all the 9 testing simulations with 962 monthly time steps (**Fig. 6**). The RMSE of pressure through the layers is relatively stable, with mean level at 3.78 psia, whereas the RMSE of saturation through the layers is quite low in most of the layers except that it is relatively high (0.06) in layer 16 or below. It is challenging to predict saturation in Layer 16, because it contains the maximum number (464) of inactive grid cells making the $CO_2$ plumes extremely small and irregularly distributed (**Fig. 7**). As a result of the large number of inactive grid cells, this layer serves like a flow barrier for the layers below it (layer 17 to 20), and is likely to impact the saturation prediction quality in these layers as well. Interestingly we observe that the saturation RMSE monotonically decreases from layers 16 to 25, which effectively demonstrates the capability of the saturation DNN model in Scenario 5 as well.

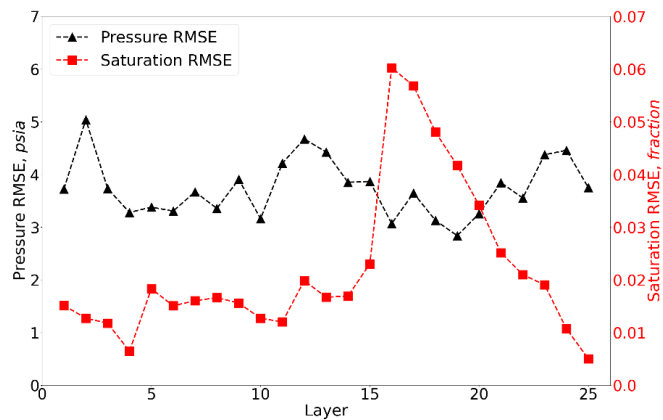

**Fig. 6. RMSE of pressure (predicted by Scenarios 1 and 3) and saturation (predicted by Scenario 5) aggregated through all the 9 testing simulation runs and 962 time steps.**

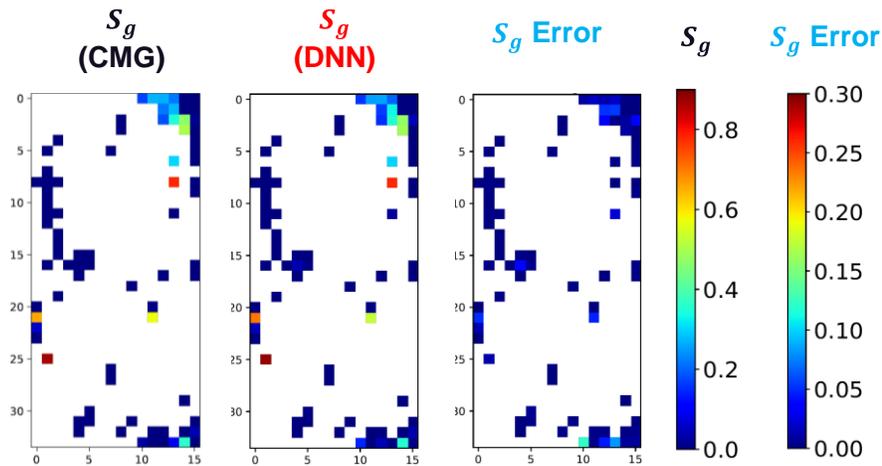

**Fig. 7. $CO_2$ saturation in layer 16 (white: inactive grid cells) at 80 years in testing simulation run 82**

Further, we select a representative testing simulation run (Run 82), and visualize its pressure and saturation fields in the Layers 1 (top), 14 (middle) and 25 (bottom) at years 1.56 (early injection), 15.58 (middle injection), 31.50 (early post-injection) and 79.53 (close to the end of post-injection). In order to demonstrate the capability of time-interpolation for our DNN models, **Fig. 8** presents the selected 4 testing time steps (blue stars), which are actually not available in the original training time steps (red dots). **Fig. 8** also shows that the training time steps are evenly spaced with yearly resolution except the monthly resolution refined in the 1$^{st}$ year and 30$^{th}$ year when injection and post-injection just start, respectively.

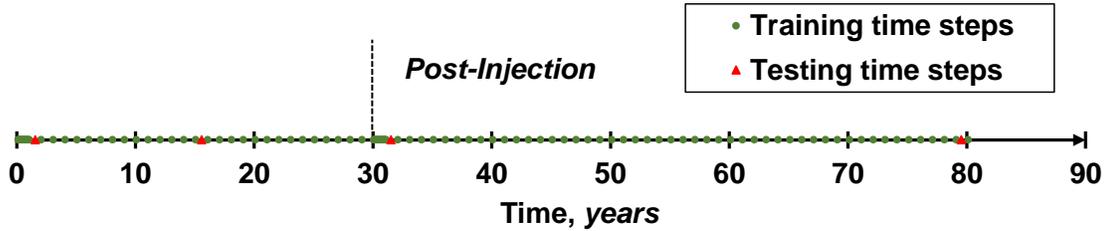

**Fig. 8. Select 4 testing time steps that are not available in the training time steps**

**Fig. 9** and **10** illustrate the pressure and saturation comparison between reservoir simulation and DNN predictions in the injection period (years 1.56 and 15.58). The DNN prediction of pressure is quite accurate, with mean absolute errors (MAE) 2.79 psia and 3.35 psia in these two interpolated time steps compared to CMG reservoir simulation. With continuous $CO_2$ injection and water production in this period, the pressure changes globally. The water production also creates low pressure regions nearby the water wells, as shown in the green pressure regions in the center of Layer 25. Further, the rock heterogeneity also impacts the pressure patterns, for example, the low pressure zone (blue) in Layer 1. These patterns in pressure fields are accurately honored by our DNN model. As $CO_2$ is injected into the reservoir, the $CO_2$ plumes in each layer grow surrounding the injection wells with time (e.g. Layer 1), and the $CO_2$ plume size decreases with increasing depth because of the buoyancy effect. These details are also accurately predicted by our DNN model, with the saturation MAE 2.07e-3 and 3.52e-3 in the two injection time steps, respectively.

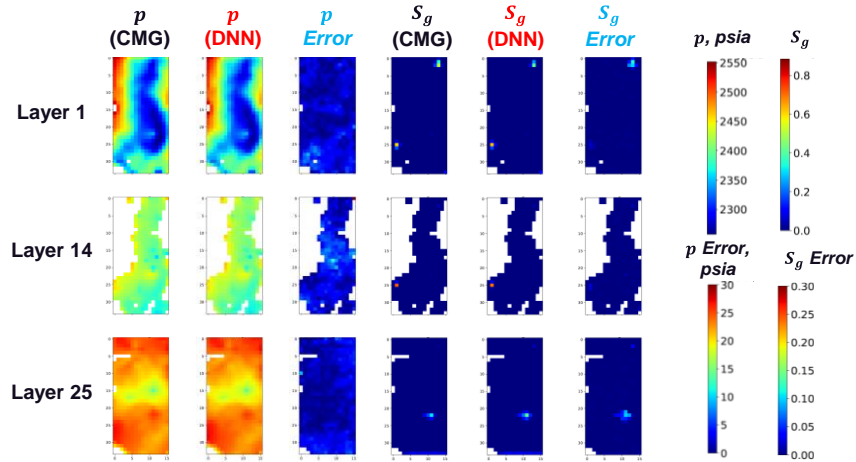

**Fig. 9. Pressure and Saturation maps of Run 82 at 1.56 years (early injection). Pressure by DNN Scenario 1, with MAE 2.79 psia; saturation by DNN Scenario 5, with MAE 2.07e-03.**

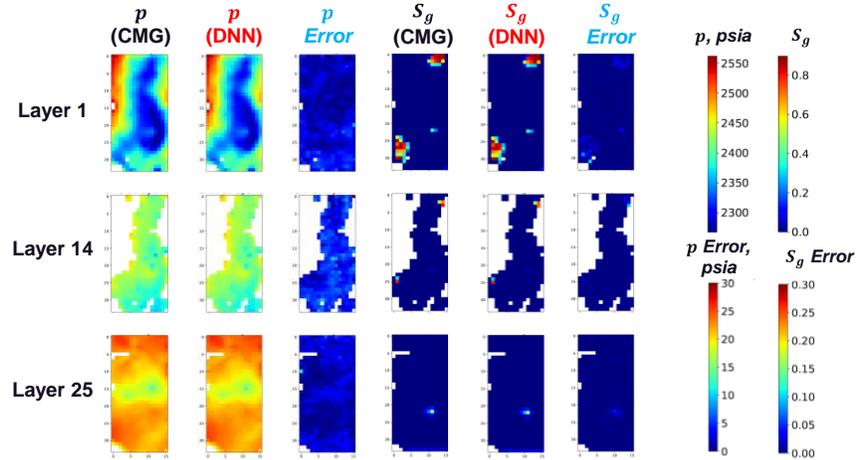

**Fig. 10. Pressure and Saturation maps of Run 82 at 15.58 years (middle injection). Pressure by DNN Scenario 1, with MAE 3.35 psia; saturation by DNN Scenario 5, with MAE 3.52e-03.**

**Fig. 11** and **12** presents the pressure and saturation predictions in the post-injection period (years 31.50 and 79.53). With all wells shut-in in this period, the fluids in the reservoir starts re-equilibrium driven by gravity. We clearly observe the disappearance of the low pressure zone nearby the production wells in the center of Layer 25, which shows during the injection period. Besides, the pressure at the Layer 25 at 79.53 years in **Fig. 12** significantly builds up (dark red) after water slowly accumulates at the bottom of the reservoir because it is heavier than $CO_2$. DNN captures these fine details, and the MAEs of the two snapshots are extremely low, 2.05 and 1.65 psia, respectively. As $CO_2$ injection ceases, the $CO_2$ mass stored in the reservoir becomes constant and the plume sizes grow only slightly (e.g., Layer 1). This is well captured by our DNN model, with the mean absolute errors 4.32e-3 and 5.50e-3 in the two post-injection time steps, respectively.

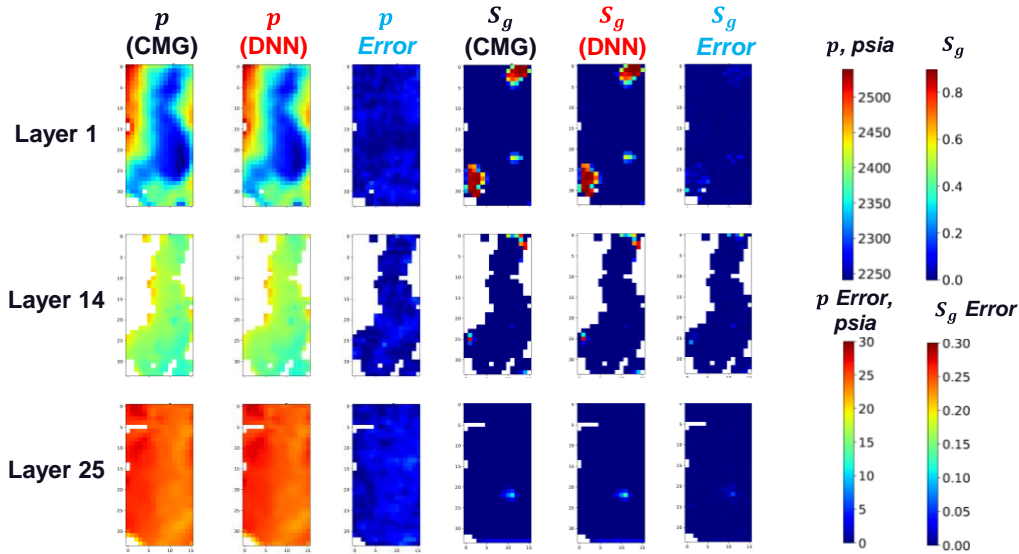

**Fig. 11. Pressure and Saturation maps of Run 82 at 31.50 years (early post-injection). Pressure by DNN Scenario 3, with MAE 2.05 psia; saturation by DNN Scenario 5, with MAE 4.32e-03.**

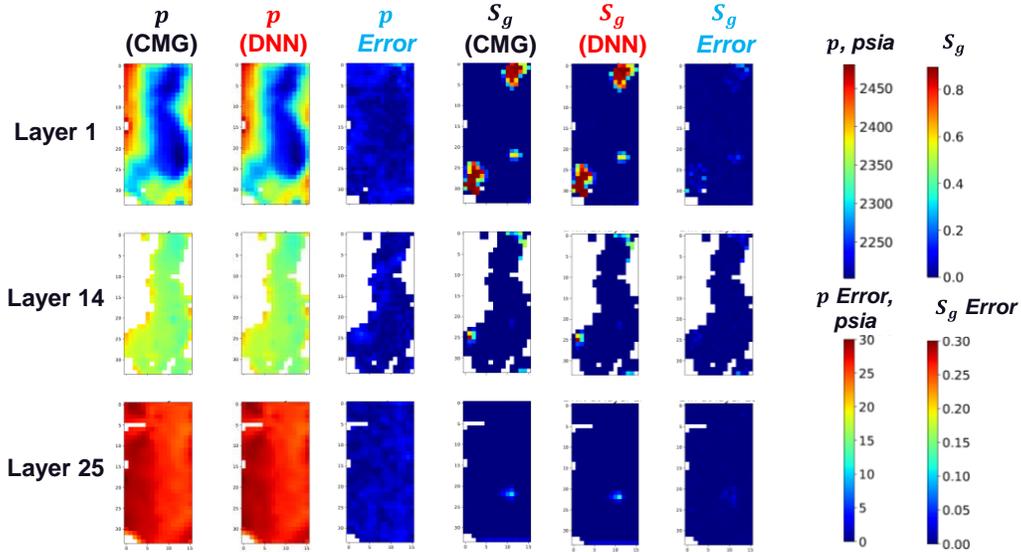

**Fig. 12. Pressure and Saturation maps of Run 82 at 79.53 years (end post-injection). Pressure by DNN Scenario 3, with MAE 1.65 psia; saturation by DNN Scenario 5, with MAE 5.50e-03.**

In **Fig. 13 (a)** to **(e)**, we also plot the pressure prediction with time at the 5 well grid cells (potential pressure monitoring locations) at layer 1, and our DNN predictions are well aligned with the CMG simulations, with the highest error only 0.12% among them. Besides, there is less pressure dynamics in the injection period as $CO_2$ injection and water production competes simultaneously in the reservoir. However, the pressure significantly declines and ultimately levels off in all the 5 well grid cells at layer 1 (reservoir top), because the heavier water-rich phase flows downwards while the lighter $CO_2$-rich phase buoys upwards because of their density difference. Our DNN predictions successfully honor these changes as well, which is likely due to that it is quite an appropriate strategy to train two exclusive pressure DNN models for injection and post-injection periods. Since the saturations of the $CO_2$-rich phase in the injection well grid cells are quite stable and always close to 1, we did not plot the saturation changes in these grid cells. Instead we plot the dynamic changes of the total $CO_2$ pore volume in the whole reservoir, calculated by $\sum S_g \phi$. In **Fig. 13 (f)**, the error of $\sum S_g \phi$ of DNN prediction compared to CMG simulation is 1.82%, which is still quite accurate. Besides, DNN prediction of $\sum S_g \phi$ clearly shows the linear trend in the injection period, because the injection rate of $CO_2$ is quite stable and greater than 0, and the post-injection plateau results from the shut-in of all wells.

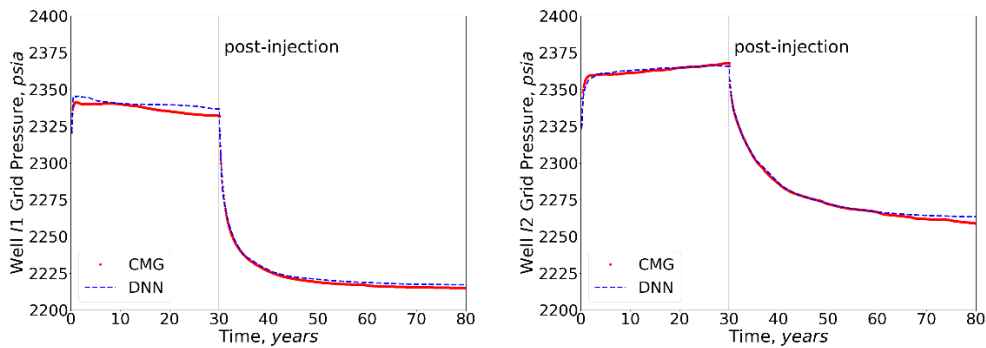

(a) Pressure at well grid (23, 12, 1), error: 0.12%; (b) Pressure at well grid (3, 14, 1), error: 0.075%

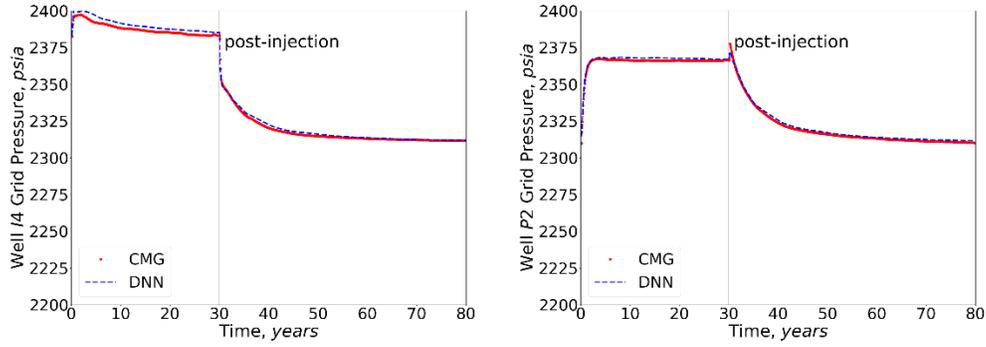

(c) Pressure at well grid (26, 2, 1), error: 0.090%; (d) Pressure at well grid (18, 6, 1), error: 0.058%

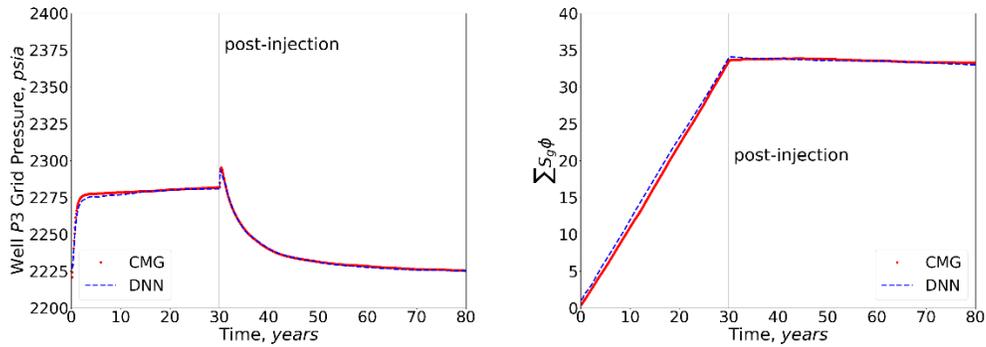

(e) Pressure at well grid (16, 12, 1), error: 0.051%; (f) Total $CO_2$ pore volume ($\sum S_g \phi$), error: 1.82%

**Fig. 13. Pressure and Saturation comparison between DNN and CMG. (Error based on $L2$ norm)**

In **Table 4** we compare the prediction CPU time of DNN versus CMG simulation when predicting the 3D pressure and saturation fields over 962 time steps (80 years). The CPU time for CMG to predict pressure and saturation is 42.18 minutes, which is expensive due to full physics compositional simulation. However, the total CPU time for DNN to predict pressure and saturation sequentially is only 0.17 minutes, which is 251 times faster than CMG. As DNN actually predicts the pressure and saturation separately, it will take only 0.087 minutes if we predict them concurrently (in parallel). Besides, for tasks such as inverse modeling (Chen et al., 2018; Chen et al., 2002) we do not need to predict the snapshots of state variable at all the monthly time steps of 80 years. Instead, we may just need to predict snapshots only in yearly resolution to match sparse observation data, and our DNN models efficiently predict the snapshots of the sparse time steps without any dependency on their previous time steps.

**Table 4. Prediction time comparison per simulation run with 962 time steps**

|  | DNN Prediction CPU, minutes | CMG Simulation CPU, minutes |
| --- | --- | --- |
| Pressure (Scenario 1 & 3) | 0.087 | NA |
| Saturation (Scenario 5) | 0.081 | NA |
| Total | 0.17 | 42.18 |

## 5. Discussions and Conclusions

In this work, a deep learning (DL) workflow based on Fourier Neuron Operator is developed to learn the complex flow physics in long term process of geologic $CO_2$ sequestration from reservoir simulation data, and is capable of accurately inferring the temporal-spatial evolution of pressure and $CO_2$ plumes. To improve the prediction fidelity, we comprehensively evaluate different sets of features that are fed into the DL workflow, and carefully examine the necessity to train two separate or a single predictive model for the injection period and the following post-injection period, as the primary driving force of fluid flow and transport in these two periods is different.

We successfully applied the DL workflow into a realistic example of $CO_2$ injection and storage in a 3D heterogeneous saline aquifer, and demonstrated that two deep neural network (DNN) models to separately predict the injection and post-injection periods provide the most accurate pressure predictions, with RMSE 3.78 psia, and also lead to minimum spikes at the early periods of injection and post-injection when transient flow is significant. Besides, in the post-injection period it is important to use cumulative $CO_2$ injection volume $Q_{inj}$ to inform the DNN about the total storage amount of $CO_2$. As for saturation, a single DNN model that takes $Q_{inj}$ as its feature is sufficient to accurately predict both injection and post-injection periods, with RMSE 0.0254. As the reservoir simulation output of pressure and saturation has a lot of time steps (962) in monthly resolution, we successfully reduced the training and validation samples with only 104 time steps through strategically selecting year-resolution time steps except month-resolution time steps in the 1$^{st}$ year of injection and post-injection periods. This approach successfully reduces the sample size by 89% and makes the training process quite affordable, with total training time of the DNN models 7.11 hours. Besides, we did not observe any loss of fidelity in pressure and saturation when we predict these intermediate time steps that are not considered during the training process. We also gauge the dynamic pressure prediction at the 5 well grid cells, which can be potential monitoring locations, and the DL workflow exhibit high accuracy with a maximum error of only 0.12%. The total prediction time of the DL workflow is about 0.017 minutes (or 10.2 seconds), which is 251 times faster than the full physics simulation based on CMG simulator (42 minutes). The efficiency and accuracy of the DL workflow across the time and spatial scales makes it a robust prediction tool for long term GCS processes, and greatly favors tasks of inverse modeling that require hundreds or thousands of forward simulation.

**Acknowledgement**

The authors acknowledge the financial support by US DOE's Fossil Energy Program Office through the project, Science-informed Machine Learning to Accelerate Real Time (SMART) Decisions in Subsurface Applications. Funding for SMART is managed by the National Energy Technology Laboratory (NETL). The authors also thank Dr. Wei Jia from University of Utah for providing the reservoir simulation data for geologic $CO_2$ sequestration, and thank Dr. Diana Bacon from Pacific Northwest National Laboratory for providing parsing tools to process the simulation data.